\documentclass{article}

\usepackage{arxiv}

\usepackage[utf8]{inputenc} 
\usepackage[T1]{fontenc}    
\usepackage{hyperref}       
\usepackage{url}            
\usepackage{booktabs}       
\usepackage{amsfonts}       
\usepackage{nicefrac}       
\usepackage{microtype}      
\usepackage{lipsum}
\usepackage{graphicx}
\graphicspath{ {./images/} }

\usepackage{algorithm,algorithmic}%
\usepackage{multirow}
\usepackage{amsmath}
\DeclareMathOperator*{\argmax}{arg\,max}

\title{A system of vision sensor based deep neural networks for complex driving scene analysis in support of crash risk assessment and prevention}

\author{
 Muhammad Monjurul Karim \\
  Department of Civil Engineering\\
  Stony Brook University\\
  Stony Brook, NY 11794, USA \\
  \texttt{muhammadmonjur.karim@stonybrook.edu} \\
   \And
 Yu Li \\
  Department of Civil Engineering\\
  Stony Brook University\\
  Stony Brook, NY 11794, USA \\
  \texttt{yu.li.5@stonybrook.edu} \\
  \And
 Ruwen Qin \\
  Department of Civil Engineering\\
  Stony Brook University\\
  Stony Brook, NY 11794, USA \\
  \texttt{ruwen.qin@stonybrook.edu} \\
  \And
 Zhaozheng Yin \\
  Department of Computer Science\\
  Stony Brook University\\
  Stony Brook, NY 11794, USA \\
  \texttt{zyin@cs.stonybrook.edu} \\
}

\begin{document}
\maketitle
\begin{abstract}
To assist human drivers and autonomous vehicles in assessing crash risks, driving scene analysis using dash cameras on vehicles and deep learning algorithms is of paramount importance. Although these technologies are increasingly available, driving scene analysis for this purpose still remains a challenge. This is mainly due to the lack of annotated large image datasets for analyzing crash risk indicators and crash likelihood, and the lack of an effective method to extract lots of required information from complex driving scenes. To fill the gap, this paper develops a scene analysis system. The Multi-Net of the system includes two multi-task neural networks that perform scene classification to provide four labels for each scene. The DeepLab v3 and YOLO v3 are combined by the system to detect and locate risky pedestrians and the nearest vehicles. All identified information can provide the situational awareness to autonomous vehicles or human drivers for identifying crash risks from the surrounding traffic. To address the scarcity of annotated image datasets for studying traffic crashes, two completely new datasets have been developed by this paper and made available to the public, which were proved to be effective in training the proposed deep neural networks. The paper further evaluates the performance of the Multi-Net and the efficiency of the developed system. Comprehensive scene analysis is further illustrated with representative examples. Results demonstrate the effectiveness of the developed system and datasets for driving scene analysis, and their supportiveness for crash risk assessment and crash prevention.
\end{abstract}


\section{Introduction}
Driving scene analysis is an emergent topic of both driver assistance and autonomous driving technology. It provides a driver or an autonomous vehicle the situational awareness of the surrounding traffic. For the purpose of crash risk assessment and crash prevention, driving scenes are complex because multiple elements jointly contributing or relating to crashes. These include, but not limited to, abnormality, nearby vehicles and pedestrians, road type and configuration, weather, and visibility. Therefore, a comprehensive analysis of driving scenes is a prerequisite for a reliable prediction of the immediate future risk condition. Multiple tasks are involved to deliver a comprehensive result of scene analysis, including object detection, instance segmentation, and classifications of scenario, road type, weather, visibility, and so on. A straightforward strategy is to have a system of different tools with each being dedicated to one or a few sub-tasks of driving scene analysis.

Increased availability of dash cameras for vehicles are adding the visual sensing capability for driving scene analysis. Meanwhile, deep learning methods such as Convolutional Neural Networks (CNNs) have achieved tremendous success in various computer vision tasks such as image classification \cite{rawat2017deep}, object detection \cite{ zhao2019object}, and instance segmentation \cite{ garcia2017review}. CNN has many potentials for driving scene analysis. Yet, driving scene analysis in support of crash risk assessment and crash prevention has not been a thoroughly solved problem, probably due to the lacks of effective methods of complex scene classification, a guidance for assessing crash risks from scene analysis results, and publicly available datasets that capture risk indicators and crash likelihood from complex driving scenes.

To tackle these challenges, various sub-tasks for driving scene analysis were developed, such as vehicle detection \cite{ hu2019joint}, pedestrian detection \cite{ liu2019high}, lane finding \cite{ hou2019learning}, traffic sign detection \cite{ zhu2016traffic}, weather recognition \cite{ zhao2018cnn}, and infrastructure and traffic recognition \cite{sikiric2014multi}. A few exceptions are noticed, which can describe traffic scenes with multi-label classification \cite{ chen2019deep}. However, the work of \cite{ chen2019deep} is solely an image classification problem that completely relies on image-centric features for scene classification. Scene classification by crash likelihood and crash risk indicators is essential for crash prevention, but no work has thoroughly studied this problem. Researchers have also been trying to develop large-scale datasets \cite{ deng2009imagenet, lin2014microsoft, zhou2017places} and deep learning models \cite{ ren2015faster, redmon2018yolov3} that well support classification tasks for understanding natural scenes. Some labeled traffic-scene datasets have been developed too \cite{ geiger2012we, cordts2016cityscapes, ramanishka2018toward, yu2018bdd100k} to address the need for traffic scene understanding. These datasets are mainly focused on environmental perception and semantic segmentation. Above-mentioned efforts are still insufficient to deliver a satisfying  driving scene analysis result to support crash risk assessment, because lacking an ability to capture deep features required to explain crash risks, such as crash likelihood or road function. Being able to identify the pre-crash scenario would allow for taking an immediate action of crash prevention. Crashes have varied features across different road types. Classifying driving scenes by road function (i.e., interstate, collector, local, and so on) can provide important clue for assessing crash risks \cite{Li2021crash}.

To fill some of the identified gaps, this paper developed a system for complex driving scene analysis in support of crash risk assessment and crash prevention. Firstly, this paper developed two new annotated image datasets to support the crash scene classification (no crash, pre-crash, crash) and road function classification, respectively. Secondly, it built a system of new multi-task neural networks named Multi-Net to generate multiple labels for classifying driving scenes. Thirdly, the Multi-Net was integrated with an object detector Yolo v3 \cite{ redmon2018yolov3} and an instance segmentation tool DeepLab v3 \cite{ chen2017rethinking} to characterize vehicles and pedestrians that may cause a crash.

The remainder of this paper is organized as the following. The next section delineates the method to develop the proposed system, followed by the description of the developed datasets. The following section presents an evaluation of the developed system illustrates various examples of scene analysis results. The conclusion and future work are summarized at the end, in the last section.
\begin{figure}[!ht]
  \centering
  \includegraphics[width=0.9\textwidth]{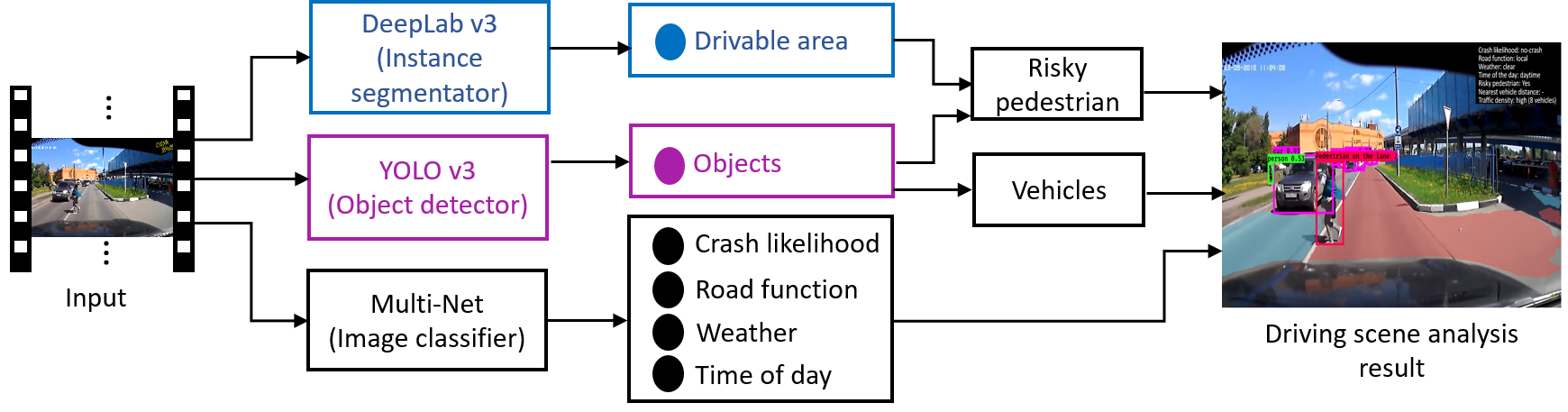}
  \caption{The schematic diagram of the driving scene analysis system} \label{fig:overview}
\end{figure}

\section{The system for complex driving scene analysis}
\label{section:multi-net}
The proposed system for driving scene analysis is illustrated in FIGURE \ref{fig:overview}. Driving scenes, in the form of RGB images captured by dash cameras mounted on vehicles, will be processed to identify the following information of each frame: the drivable area of the road, vehicles, risky pedestrians, crash likelihood, road function, weather, time of day. These information are critical for assessing crash risks \cite{Li2021crash}. The deep learning models for analyzing driving scenes include the multi-task deep neural networks (Multi-Net),  an object detector YOLO v3 \cite{redmon2018yolov3}, and an image segmentation algorithm DeepLab v3 \cite{chen2017rethinking}. The Multi-Net performs multi-task image classifications to provide the labels of crash likelihood, road function, weather, and time of day for driving scenes. YOLO v3 detects objects in video frames, such as pedestrians and vehicles. The distance to the nearest vehicle is calculated in every frame. The DeepLab v3 segments drivable area. Risky pedestrians who are within the drivable area are also detected.

\subsection{Multi-Net for multi-task classification of driving scenes}
\begin{figure}[!ht]
  \centering
  \includegraphics[width=1\textwidth]{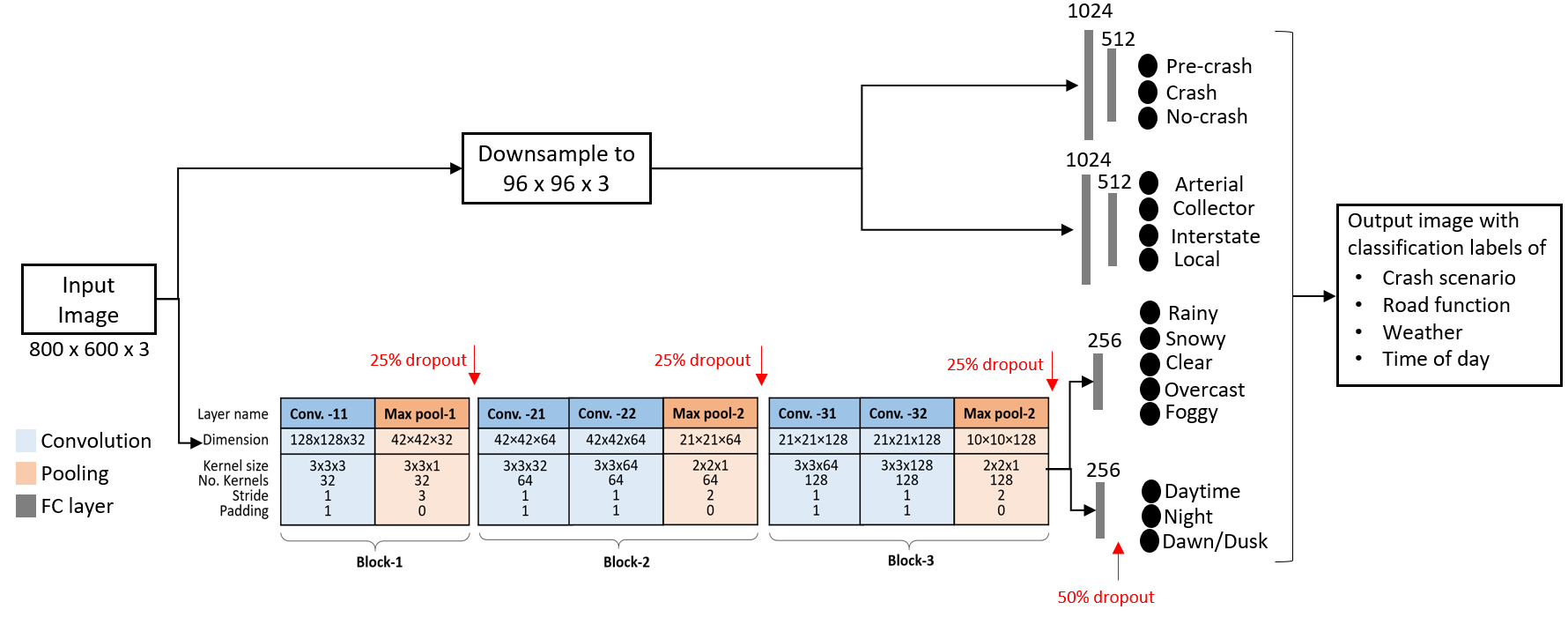}
  \caption{The architecture of the Multi-Net.}\label{fig:multi-net}
\end{figure}

A parallel neural network model, named Multi-Net and shown in FIGURE \ref{fig:multi-net}, was developed in this study to recognize the crash likelihood and another three crash risk indicators - road function, weather, and time of day. The Multi-Net comprises of two parallel multi-task networks and each network is further split into two branches. Thus, the Multi-Net has four branches in total, which respectively perform four classification tasks to determine the class of the four variables of driving scenes. For example, the crash likelihood has three possible classes: pre-crash, crash, and no crash. The classes of road function, weather, and time of the day are displayed in FIGURE \ref{fig:multi-net}. 

Input RGB images to the Multi-Net are of size $w\times h \times c$, where $w$, $h$, and $c$ are the width, height, and the number of channels of the images. The first network downsamples each input image to $96\times 96\times 3$, flats the downsampled image, and then feeds it to two separate multi-level perceptron branches for classifying crash likelihood and road function, respectively. Each perceptron branch has two hidden layers. The first layer has 1024 neurons and the following layer has 512 neurons. Then, the final layers of the two branches respectively generate the class scores for crash likelihood and road function.

The second network is a Convolutional Neural Network (CNN) for classifying both weather and time of day. Since almost all the pixels of an image contains information of these two variables, a CNN is more capable than a perceptron in capturing the information, particularly the spatial dependency in an image, through the convolution and pooling operations. Firstly, input images to this CNN are downsampled to $128\times 128\times 3$. Then each downsampled image is processed by three blocks of convolutional layers and pooling layer. The kernels, stride, and padding of convolution and pooling operations are delineated in FIGURE \ref{fig:multi-net}.
The convolutional layers are the feature extractor of the CNN. A feature map of size $10\times 10\times 128$ is created from the last convolution layer. The feature map is sent to two branches of a fully connected layer with 256 neurons, which respectively perform the classifications of weather and time of day to generate class scores.

\subsection{Object detection with YOLO v3}
YOLO v3 is an object detector that can provide real-time object detection. It is a single-stage detection network that takes an image as the input and detects objects in the image by predicting their bounding box coordinates, the confidence of the bounding boxes, and class probabilities. Compared to its ancestors, YOLO v3 has a better performance in detecting small objects. Interested readers can refer to \cite{redmon2018yolov3} for details.

\subsection{Distance to the nearest vehicle}
\label{subsection:distance}
\begin{figure}[!ht]
  \centering
  \includegraphics[width=0.85\textwidth]{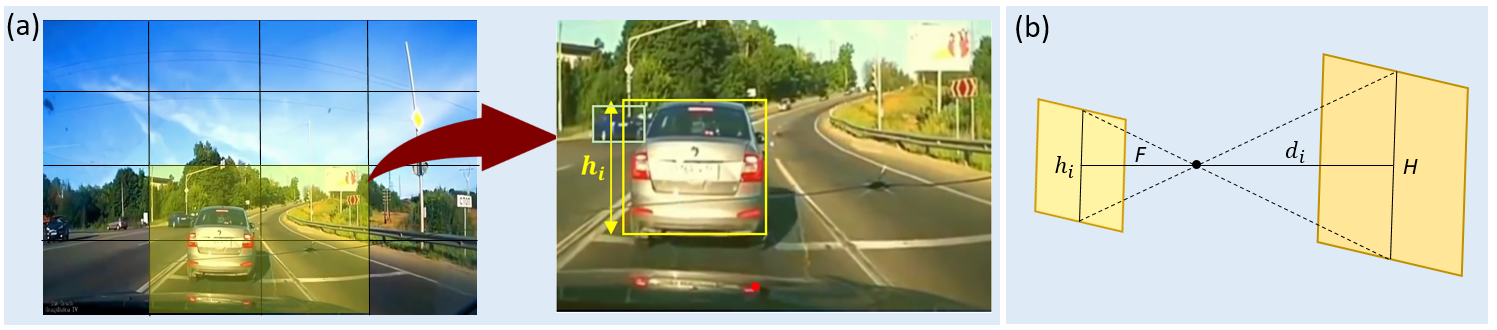}
  \caption{Measuring the distance to another vehicle.}\label{fig:vehicle_distance}
\end{figure}

The distance to the nearest vehicle along the traveling path is an important measurement for crash risk assessment. This study divided any input image into a 4$\times$4 grid and concentrated on the area containing the travel path of the user's vehicle to reduce computational cost. FIGURE \ref{fig:vehicle_distance}(a) illustrates an example wherein the traveling path of the user's vehicle is most likely found in the portion highlighted in yellow. Only vehicles in this highlighted region of interest are considered for the distance calculation.

Assume the region of interest contains $N$ objects indexed by $i$. Let $h_i$ be the height of vehicle $i$ in the image and $H$ be the true height. Let the focal length of the dash camera be $F$ and $d_i$ be the distance from the camera to vehicle $i$. Accordingly to the principle of similar triangles, $d_i/H=F/h_i$. 
This study assumes that $F$ is 2.5 inch, $H$ is 7 foot for a Van, 6 foot for an SUV, and 4.7 foot for a car. $h_i$ is measured in pixel, and pixels per inch is 100. Therefore, the distance to vehicle $i$ is
\begin{equation}
    d_i=HF/h_i=250H/h_i,
 \end{equation}
and the nearest vehicle is identified as
\begin{equation}
    \argmax_i(250H/h_i).
\end{equation}

\subsection{Detecting drivable area and risky pedestrian}

\begin{figure}[!ht]
  \centering
  \includegraphics[width=0.8\textwidth]{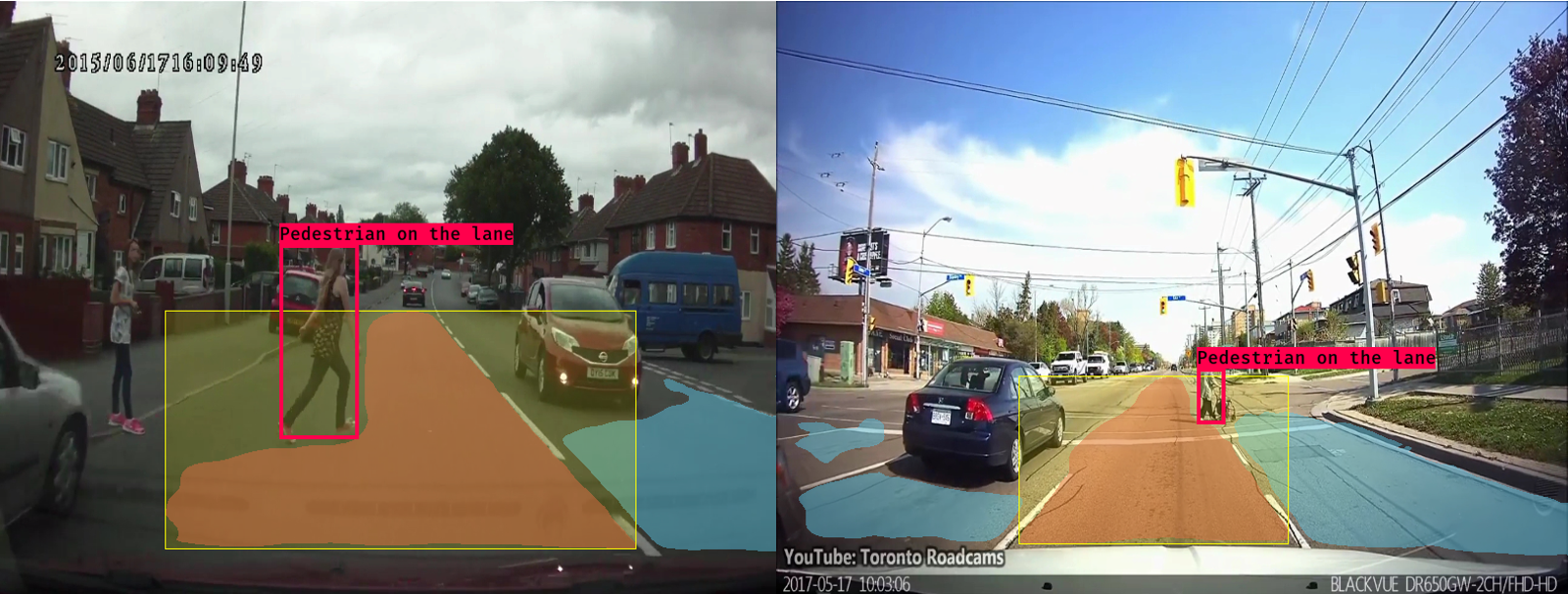}
  \caption{Detecting drivable area and risky pedestrians.}\label{fig:risky_pedestrian}
\end{figure}

The ability to segment different areas related to trafficway (such as on the trafficway, shoulder, roadside, median, and so on) is useful for crash risk assessment because crash risks differ in those areas. For example, pedestrians near the user's vehicle, particularly those on the same lane as the user, maybe risky and require an attention. Therefore, this study created the capability of detecting risky pedestrians. DeepLab v3 trained with the BDD100k dataset \cite{yu2018bdd100k} was used in this study to segment the drivable area that includes two classes of lanes, the direct lane where the user is currently driving and the alternative lanes where the user can go by lane changing. In FIGURE \ref{fig:risky_pedestrian}, lanes highlighted in red color are direct lanes and those in blue are alternative lanes.

DeepLab v3 cannot segment the area where objects like pedestrians and vehicles are present. Therefore, DeepLab by itself cannot identify the relationship of pedestrians to the driving lane. This study proposed an algorithm of risky pedestrian detection below. A rectangle bounding box $b_L$ is extrapolated using the most outside coordinates of the segmented direct lane.
If the YOLO v3 object detector detects $P$ pedestrians in the driving scene, indexed by $p$, it returns all possible bounding boxes $b_{p}$'s. If the two bounding boxes $b_L$ and $b_p$ have an overlap, pedestrian $p$ is a risky one and a notification of risk is generated and the distance to that pedestrian is similarly estimated using the same method introduced in this paper to estimate the distance to the nearest vehicle. In FIGURE \ref{fig:risky_pedestrian}, a yellow bounding box represents the extrapolated bounding box of the direct lane, and the red one represents the bounding box for the pedestrian. In both frames of FIGURE \ref{fig:risky_pedestrian}, these two boxes overlap each other, indicating the pedestrian is in the direct lane. 

\begin{algorithm}[!ht]
\caption{Risky pedestrian detection}\label{algorithm_2}
\begin{algorithmic}
\STATE {$S_L \gets \{(x_1,y_1 ),(x_2,y_2 ),…,(x_n,y_n )\}$}; \COMMENT{$S_L$ is the segmented polygon for direct lane}
\STATE {$b_L \gets \{(x_{min},y_{min} ),(x_{min},y_{max} ),(x_{max},y_{min} ),(x_{max},y_{max} ) \}$}; \COMMENT{$b_L$ is the minimum rectangle containing $S_L$}
\FOR{$p \gets 1\ to \ P$}
\STATE {$b_{p} \gets \{(x_{1},y_{1} ),(x_{2},y_{2} ),(x_{3},y_{3} ),(x_{4},y_{4} ) \}_{p}$}; \COMMENT{bounding box for pedestrian $p$}
\IF{$b_L \cap \ b_{p} \neq \emptyset$ }
\STATE {Generate a risk notification.}
\ENDIF
\ENDFOR
\end{algorithmic}
\end{algorithm}


\section{DATASET DEVELOPMENT}
\label{section:dataset}
\subsection{Data acquisition}
\begin{figure}[!ht]
  \centering
  \includegraphics[width=0.85\textwidth]{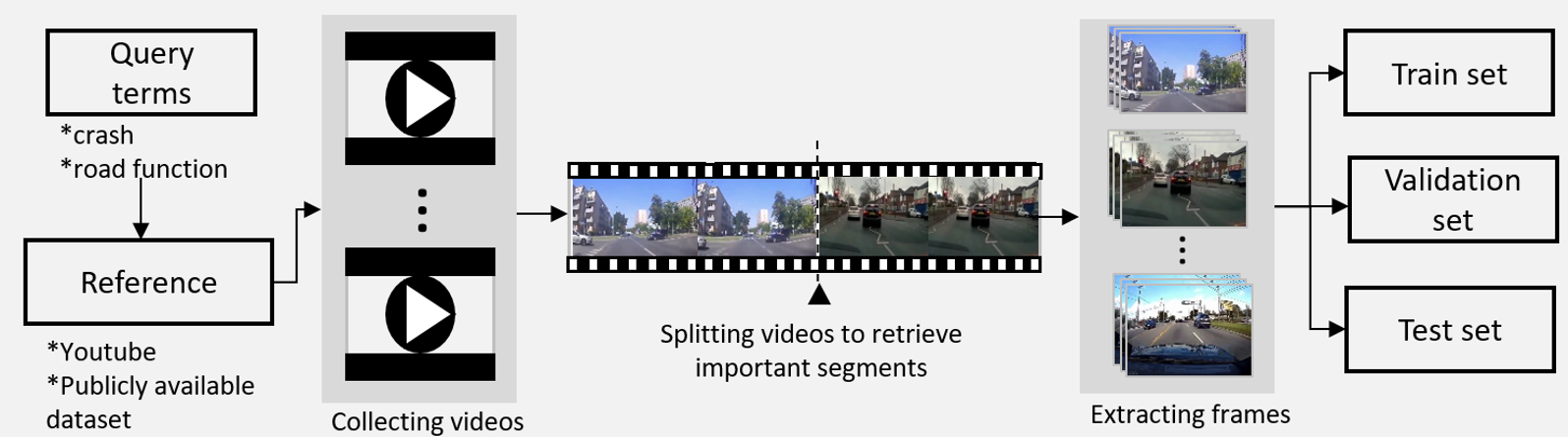}
  \caption{The pipeline for mining images for dataset development.}\label{fig:dataset_development}
\end{figure}

No public datasets are available for training a deep learning neural network to classify crash likelihood with a pre-crash class and to classify the US road functional system. This study addressed this issue by developing two datasets that contain a diverse set of driving scenes with various road types and crash scenarios. The method for dataset development is illustrated in FIGURE \ref{fig:dataset_development}. This study collected crash videos from YouTube and general driving scene videos from HDD \cite{ramanishka2018toward} using query terms like ``crash'' and ``road function''. Segments containing classes of interests were retrieved, which were further converted into image frames. Consecutive frames of each video segment are similar and, therefore, sampling a portion of frames that are evenly distributed on the timeline would be sufficient to represent the segment. This study sampled one frame and then skipped five frames. The sampled frames were randomly split into training, validation, and testing sets.

The collected datasets were manually labeled with the classes of road function and crash likelihood. FIGURE \ref{fig:multi-net} shows four general classes of road function: 1) arterial, 2) collector, 3) interstate, and 4) local per the road functional system definition from \cite{federal2013highway}. Crash likelihood has three classes. When a vehicle collides with other objects, then the scenario is defined as a crash. A clip of length two seconds before the crash is defined as pre-crash. Three seconds before the pre-crash is defined as no-crash.

\subsection{Description of the datasets}

TABLE \ref{tab:dataset} below summarizes the datasets, which can be accessed at \cite{crash_data_2020}. Each dataset is split into three mutually exclusive subsets: training, validating, and testing sets. In each dataset, samples are approximately evenly distributed among classes. 

\begin{table}[!ht]
	\caption{Summary of the datasets}\label{tab:dataset}
	\begin{center}
		\begin{tabular}{l| c| r |r r r}
			Task & No. of Classes & Total & Train & Val & Test \\\hline
			Crash likelihood   & 3 & 15,900 & 12,000 & 3,000 & 900\\
			Road function   & 4 & 6,400 & 4,000 & 1,200 & 1,200\\
			Weather   & 5 & 7,900 & 4,875 & 1,625 & 1,400 \\
			Time of day   & 3 & 7,400 & 4,875 & 1,625 & 900 \\\hline
		\end{tabular}
	\end{center}
	\label{tab:dataset}
\end{table}

The crash likelihood classification is challenging because of the large variety of crash scenarios. The crash dataset developed by this study contains 15,900 images.

The classification of road function is less challenging compared to crash likelihood classification because the within-class variation of road function is relatively small. Therefore, a dataset with 6,400 images was developed by this study for classifying road function. 

The dataset for weather classification was created by taking images of clear, foggy, overcast, rainy, and snowy classes from BDD100k. However, there are only 143 images of the foggy class in BDD100k. Therefore, this study generated synthetic images of the foggy class and combined these with the 143 images from BDD100k to have approximately the same amount of samples for every class of weather. Hence, this dataset has a total of 7,900 images and 6,500 of these are used for training and validation. The same training and validation sets were used for classifying time of day to leverage the annotation of the dataset.


\section{Experimental Evaluation}
\label{section:evalutation}
\subsection{Training the Multi-Net and DeepLab v3}
This study requires two training procedures, one for training the Multi-Net for multi-task classification of driving scenes and the other for training the DeepLab v3 for segmenting drivable area. The Multi-Net was trained with the dataset developed by this project on a workstation with the following configuration: a 2.90 GHz Intel Xeon W-2102 CPU with 4 CPU cores, 16GB of RAM, and an Nvidia Geforce GTX 1080 Ti GPU with 11 GB memory. The multi-task network for classifying crash likelihood and road function system was trained for 80 epochs, whereas the network for classifying weather and time of day was trained for 30 epochs. Both networks were trained using a learning rate 0.01, a batch size 32, the stochastic gradient descent optimizer, and the categorical cross-entropy loss function. Training the two perceptron branches of the Multi-Net took 32 minutes and 14 minutes, respectively. Training the CNN of the Multi-Net took 90 minutes.

DeepLab v3 was trained with the BDD100k dataset on an Intel ® Xeon ® Gold 6152 CPU @ 2.10 GHz server with 503 GB of RAM and an Nvidia Tesla V100 GPU with 32 GB of memory. This training process took place for 20 epochs with a learning rate of 0.001 and a batch size of 4 images. It took approximately 68 hours to train the DeepLab v3.

\subsection{Evaluation of classifier performance}
Three standard metrics were used for evaluating the performance of the developed Multi-net, which are precision (Pr), recall (Rc), and f1-score (F1) calculated below:

\begin{equation}
\text{Pr} = \frac{\text{No. correct predictions}}{\text{No. predictions}},\;
    \text{Rc} = \frac{\text{No. correct predictions}}{\text{No. ground-truth objects}},\; \text{F1} = \frac{2}{\text{Pr}^{-1}+\text{Rc}^{-1}}\\
\end{equation}

\begin{figure}[!ht]
  \centering
  \includegraphics[width=0.8\textwidth]{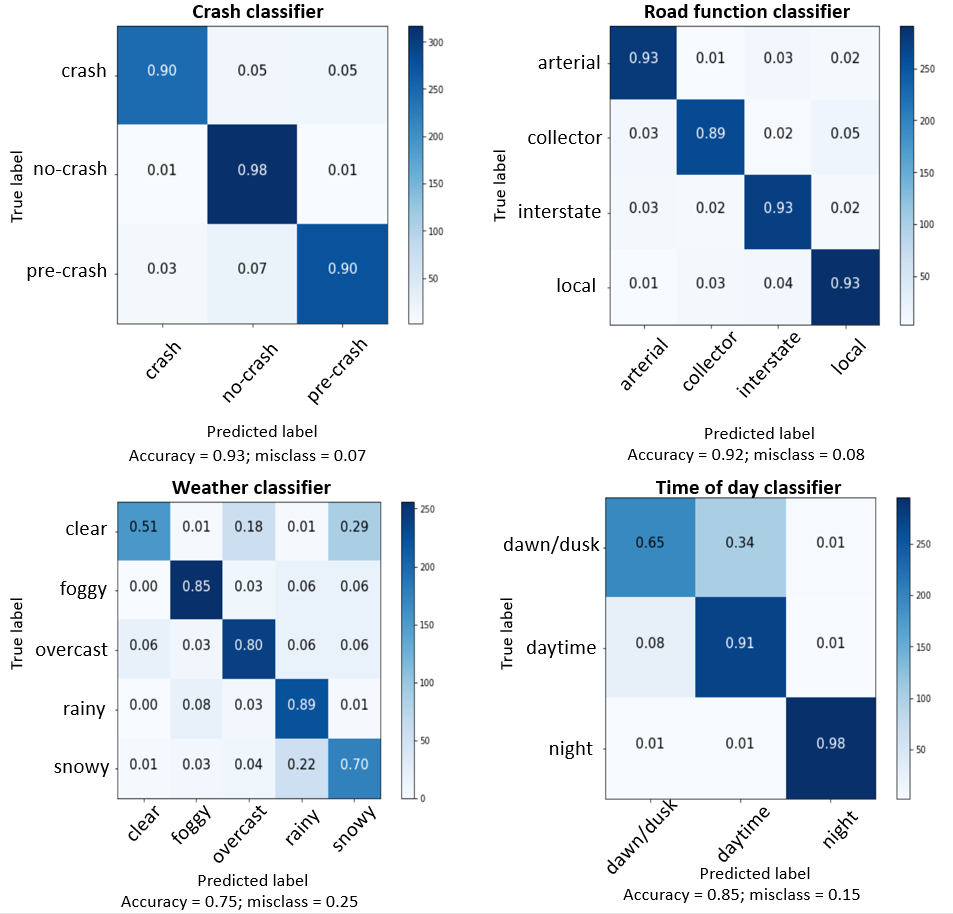}
  \caption{Confusion matrices.}\label{fig:confusion_matrix}
\end{figure}

Each classifier of the Multi-Net was evaluated on the testing dataset for it, and the performance measurements are summarized in TABLE \ref{tab:result}. The right most column shows the number of supporting images for the evaluation. The precision, recall, and f1 were calculated for each individual class of tested images. Then, the macro-average performance was calculated for the classifier, which is the average of class level performances. FIGURE \ref{fig:confusion_matrix} further shows the confusion matrices of the four classifiers to provide additional details.

\begin{table}[!ht]
	\caption{Performance of classifiers}\label{tab:result}
	\begin{center}
		\begin{tabular}{l l| r r r| r}
			Classifier & Classes & Precision & Recall & F1 & Support \\\hline
		    \multirow{4}{*}{Crash} & crash & 0.95 & 0.90 & 0.93 & 272\\
			 & no-crash & 0.90 & 0.98 & 0.94 & 322\\
			 & pre-crash & 0.95 & 0.90 & 0.92 & 306\\
			 \cline{2-6}
			 & macro-avg & 0.93 & 0.93 & 0.93 & 900\\
			 \hline
			 \multirow{5}{*}{Road function} & arterial & 0.93 & 0.93 & 0.93 & 300\\
			 & collector & 0.94 & 0.89 & 0.91 & 292\\
			 & interstate & 0.91 & 0.93 & 0.92 & 292\\
			 & local & 0.91 & 0.93 & 0.92 & 313\\
			 \cline{2-6}
			 & macro-avg & 0.92 & 0.92 & 0.92 & 1200\\
			 \hline
            \multirow{5}{*}{Weather} & clear & 0.88 & 0.51 & 0.65 & 300\\
			 & foggy & 0.87 & 0.85 & 0.86 & 300\\
			 & overcast & 0.74 & 0.80 & 0.77 & 300\\
			 & rainy & 0.71 & 0.89 & 0.79 & 250\\
			 & snowy & 0.59 & 0.70 & 0.64 & 250\\
			 \cline{2-6}
			 & macro-avg & 0.76 & 0.75 & 0.74 & 1400\\
			 \hline
			 \multirow{4}{*}{Time of day} & dawn/dusk & 0.87 & 0.65 & 0.75 & 300\\
			 & daytime & 0.73 & 0.91 & 0.81 & 300\\
			 & night & 0.98 & 0.98 & 0.98 & 300\\
			 \cline{2-6}
			 & macro-avg & 0.86 & 0.85 & 0.85 & 900\\
            \hline
		\end{tabular}
	\end{center}
\end{table}

Results in the table confirm that the multi-task network for classifying the crash likelihood and road function has a good capability of learning non-linear complex relationships for very different tasks. The precision and recall achieved by the classifier for crash likelihood are at least 90\% for any of the three classes, yielding a macro-average f1 score of 93\%. The classifier for road function achieved a similar performance. the precision and recall are 89\% or higher for any of the four classes, yielding a macro-average f1 score of 92\%.

The classifier for weather achieves a reasonable performance in classifying the foggy, overcast, and rainy classes, with 80\% or higher recall. However, the recall values of the clear and snowy classes are 51\% and 70\%, respectively, indicating that some of these instances are classified as other classes. By looking at the confusion matrix in FIGURE \ref{fig:confusion_matrix}, 29\% testing samples of the clear class are classified as snowy, and 18\% as overcast. This is due to the fact that the clear weather has a pixel distribution similar to the distributions of overcast and snowy weather. 22\% testing samples of snowy class are classified as rainy class due to the feature similarity between them. These misclassified instances lower the precision of classifying overcast, rainy, and snowy weathers.

The classifier for time of day achieves an excellent performance in classifying night instances, evidenced by the 98\% F1. Yet, this classifier has some difficulty in differentiating dawn/dusk and daytime samples due to the strong similarity in their visual features. The recall for the daytime class is 91\% but the precision is 73\%. Meanwhile, the recall for the dawn/dusk class is 65\%. From the confusion matrix in FIGURE \ref{fig:confusion_matrix} it is observed that 34\% testing samples of the dawn/dusk class are classified as daytime, thus lowering the precision of daytime class and the recall of dawn/dusk. The precision for classifying the dawn/dusk instances is 87\% and those misclassified as the dawn/dusk class are mainly from the daytime class.

\subsection{Examples of classification results}

\begin{figure}[!ht]
  \centering
  \includegraphics[width=1\textwidth]{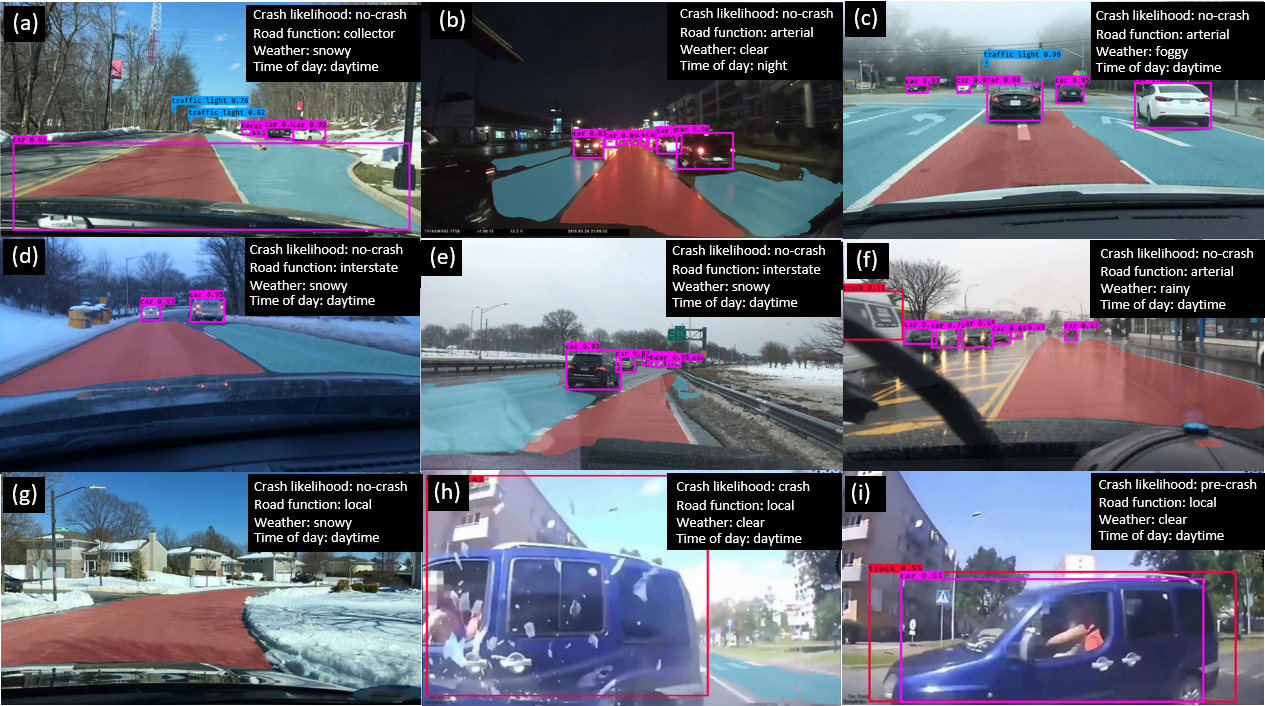}
  \caption{Examples with results generated by the Multi-net.}\label{fig:qualitative_multinet}
\end{figure}

FIGURE \ref{fig:qualitative_multinet} illustrates some examples of classification made by the Multi-Net. Four classification labels are generated for each image, shown on the top right of the image. For example, the classification result of (i) says that the driving scene is on a local road during daytime, the weather is clear, and the crash likelihood is pre-crash because the user's vehicle is about to crash into another nearby vehicle. These examples show that the Multi-Net can successfully classify the driving scenes. (h) is classified as a crash scenario when the user's vehicle collides with that vehicle, and the rest seven scenes are classified as no crash. The Multi-Net successfully classified the road function in these examples: scenes (b), (c) and (f) are classified as arterial, (a) is collector, (d) and (e) are interstate, and (g), (h) and (i) are local road. Weather can be classified too. (a), (d), (e), and (g) are classified as snowy, (b), (h) and (i) are clear, (c) is foggy, and (f) is rainy. The time of day is the fourth label of each scene. (b) is classified as night, and the rest are daytime.

\subsection{Examples of object detection}
\begin{figure}[!ht]
  \centering
  \includegraphics[width=1\textwidth]{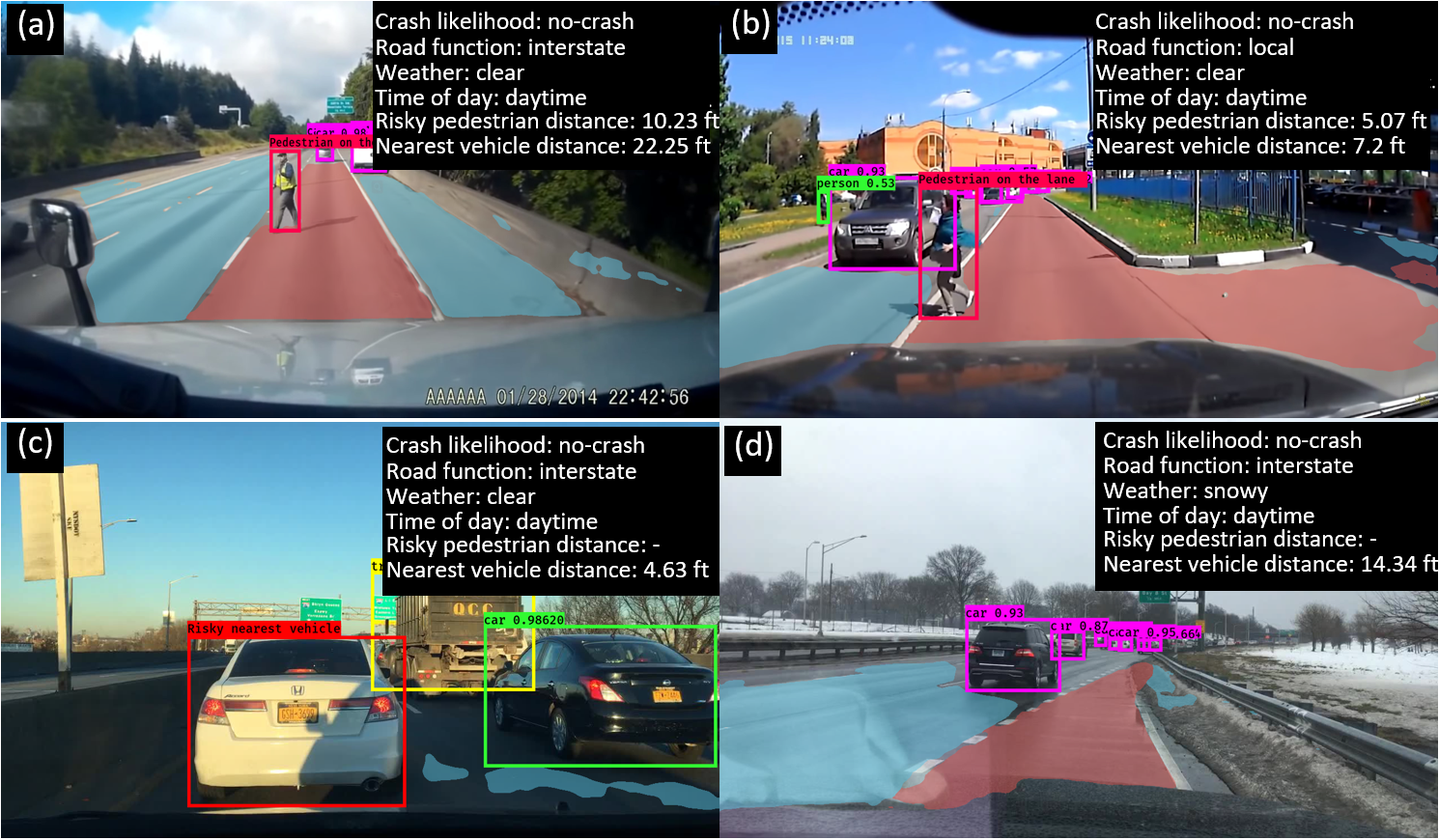}
  \caption{Example of detecting and locating risky pedestrian and the nearest vehicle. }\label{fig:additional_labels}
\end{figure}

Beyond the scene classification, the developed system detects and locates objects that may involve in a crash. The object detector YOLO v3 returns the bounding boxes of detected objects in the image. The segmentation tool DeepLab v3 segments the drivable areas with two colors: red for the direct lane and blue for the alternative lane. Based on those, the location of pedestrians and vehicles in relation to the drivable area are useful information for assessing crash risks. FIGURE \ref{fig:additional_labels} shows some examples with two additional labels. In scene (a) a pedestrian is detected on the direct lane of an interstate highway and the person is identified as a risky pedestrian highlighted with a red bounding box. In (b) two pedestrians were detected, with one on the direct lane and the other outside the drivable area. Therefore, the driving scene analysis system treats the one outside the drivable area as a safe pedestrian and highlights the person with a green bounding box. The pedestrian on the direct lane is highlighted with a red bounding box. Distance to the risky pedestrian is also calculated. For example, the risky pedestrian is 10.23 feet away from the user's vehicle in (a) and 5.07 feet away in (b). The proposed method also calculates the 
distance to the nearest vehicle. For example, in scene (c) the nearest vehicle is 4.63 feet from the user vehicle, which the user should pay attention to for avoidance of the rear-to-front collision. In the scene (d), the 14.34 feet away and it is on the alternative lane. Unlike (c), the information of the nearest vehicle in (d) does not suggest any crash risk involving another vehicle.   

\subsection{Inference speed}

The inference speed of the proposed system is critical because driving scene analysis needs to be real-time, ideally. This study thus evaluated the efficiency of the proposed system in terms of the inference speed on the workstation that was used to train the Multi-Net. The workstation is Dell Precision 5820, a typical workstation in the market. Loading the trained weights of multiple networks requires a lot of computer memory. To save the memory, the proposed system was tested in two steps. At first, testing images are passed through the DeepLab v3 for the instance segmentation. The segmented images are stored in a disk. DeepLab v3 processed  images at an inference speed of 2 frames per second (fps). Then the images were passed through Multi-Net and Yolo v3 for classification and object detection, where images were processed at a speed of 6 fps. A normal dash camera typically captures 30 fps, and this study samples 5 fps for analysis. That is, the proposed system provides a real-time inference speed without the instance segmentation. The inference speed of the segmentation task needs to be improved in order to achieve real-time scene analysis.

\section{Conclusion}
This paper presented a system for vision sensor based complex driving scene analysis in support of crash risk assessment and crash prevention. The system is composed of a Multi-Net model for scene classification, the YOLO v3 object detector, and the DeepLab v3 segmentor. Multi-Net performs scene classification and provides four labels for each scene including the likelihood of a crash, road function, weather, and time of day. The DeepLab v3 and YOLO v3 are combined to detect and locate risky pedestrians and the nearest vehicles in the scene. All these identified information can provide the situational awareness to an autonomous vehicle or a human driver for identifying crash risk from the surrounding traffic. To address the scarcity of annotated image datasets for studying crashes, two completely new datasets were developed, which were proved to be effective in training the deep neural networks with the proposed method by this study.

Vision sensor-based driving scene analysis is still facing a variety of challenges and needs. Firstly, the system can be expanded by adding the ability to classify additional risk indicators. For example, crash risks have varied features at different junction-related locations such as non-junction, four-way intersection, Y-intersection, T-intersection, driveway access, and so on. The ability to classify scenes according to their junction-related feature, as well as the ability to detect other classes of road areas such as shoulder, roadside, and median, will make the crash risk assessment more reliable. Secondly, efforts can be made to further improve the performance of the developed system. The accuracy of weather classification needs to be improved. Other methods that can substitute the segmentation method for identifying the drivable area need to be developed and integrated to the system in order to increase the inference speed. Besides, a method should be introduced to tackle the memory issue while loading trained weights from multiple networks. The estimation of the traffic density (i.e., the number of vehicles per unit length of driving lane) from dash cameras mounted on vehicles is a challenging research question but it will provide critical information for crash risk prevention. This paper has built a foundation for further exploring these exciting research opportunities.

\bibliographystyle{unsrt}  


\bibliography{references}

\end{document}